%% file: acl_latex.tex
\documentclass[11pt]{article}

\usepackage[final]{acl}

\usepackage{times}
\usepackage{amsmath}
\usepackage{amssymb}
\usepackage{latexsym}
\usepackage{graphicx}
\usepackage{booktabs}
\usepackage{algorithm}
\usepackage{algpseudocode}
\usepackage{array}
\usepackage{placeins} 
\usepackage[T1]{fontenc}

\usepackage[utf8]{inputenc}

\usepackage{microtype}

\usepackage{inconsolata}

\usepackage{tabularx}

\usepackage[most]{tcolorbox}
\usepackage{ragged2e}

\newtcolorbox{promptbox}[1]{%
  enhanced,
  breakable,
  width=\linewidth,
  colback=black!2,
  colframe=black!55,
  boxrule=0.45pt,
  arc=1pt,
  left=5pt, right=5pt, top=5pt, bottom=5pt,
  boxsep=0pt,
  fonttitle=\bfseries\small,
  fontupper=\small,
  title={#1},
  colbacktitle=black!9,
  coltitle=black,
  before upper={\setlength{\emergencystretch}{2em}},
}

\title{Principal Trait Analysis: Identifying User Behavioral Traits \\ in Human-AI Collaboration}

\author{
  Fareya Ikram\thanks{These authors contributed equally to this work.} \\
  University of Massachusetts Amherst \\
  \texttt{fikram@umass.edu}
  \And
  Hunter McNicohls\footnotemark[1] \\
  University of Massachusetts Amherst \\
  \texttt{wmcnichols@umass.edu}
  \AND
  Kai Du \\
  OpenRefinery AI \\
  \texttt{kai@openrefinery.ai}
  \And
  Andrew Lan \\
  University of Massachusetts Amherst \\
  \texttt{andrewlan@umass.edu}
}

\begin{document}
\maketitle
\begin{abstract}

Human-artificial intelligence (AI) collaboration takes many forms and occur in both professional and educational scenarios. It is important to identify how human users interact with AI and how their behavior affects task success and learning outcomes. Existing approaches often rely on predefined behavioral categories or derive patterns without establishing their relationship to outcomes. In this work, we propose Principal Trait Analysis (PTA), a data-driven method for identifying interpretable behavioral traits from human-AI collaboration dialogues. Our method analyzes large corpora of human-AI collaborative session traces, derive common traits across the dataset, scores each human collaborator on how well their interaction aligns with these traits, and selects a set that covers the most variance across users. While unsupervised and data-driven, our method also enables domain expertise to be injected during trait identification. We evaluate PTA on two human-AI collaborative coding datasets and investigate how the traits predict future task success, learning outcomes, and evolve over time. 


\end{abstract}

\section{Introduction}

LLM coding assistants are increasingly used in software development and integrated into computer science education, yet it is still unclear what distinguishes effective from ineffective use in either setting. The same model, given to two software developers, can produce vastly different code depending on how each developer interacts with the model, and the productivity gains from coding assistants vary substantially across developers~\cite{peng2023copilot}. Likewise, two students can have vastly different usage patterns which result in different learning experiences and subsequent outcomes~\cite{kazemitabaar2023,prather2024}. Identifying common usage patterns and relating them to outcomes would give i) developers guidelines on how to collaborate with coding agents and ii) instructors guidelines on how to teach ``AI use skills'' to computer science students. 

However, uncovering these patterns and establishing universal guidelines is difficult, since the space of relevant behavioral traits is large and rapidly changing as LLM capabilities improve. 
As a result, using existing theory on human cognition and behavior to define traits in a \emph{top-down} manner may quickly become stale and struggle to generalize across settings. Prior works have imported frameworks from learning theory, human-AI collaboration, or software engineering practice and applies them as fixed rubrics~\cite{ober2026}. Such rubrics are interpretable and theoretically grounded, but they are fixed at design time and can be brittle when tools and settings change, especially as AI becomes more capable at a rapid pace; measuring the competencies or skills they define is itself an open problem~\cite{zhang2026ailit}. 

To keep pace with the rapid changes in this field, more \emph{bottom-up}, automatic and data-driven analyses should be explored, so that new traits can be re-derived as human-AI collaboration evolves into new settings and scenarios.
Prior work has analyzed interaction logs directly, through qualitative coding workflows~\cite{parks2023} or dialogue-act taxonomies applied to tutoring dialogue~\cite{litman2006,graesser2016,boyer2011} and to student-AI conversations~\cite{studychat}, with more recent efforts mining such logs with LLMs in an open-ended fashion~\cite{mininggold}. These analyses stay close to observed behavior, but their annotation schemes are labor-intensive to build and maintain, and the resulting behavioral categories are only sometimes tied to measurable outcomes. Therefore, they cannot be reliably characterized as skills either: in educational research literature on knowledge components, a skill is expected to improve with practice along a learning curve and generalize across tasks~\cite{koedinger2012kli}.

\begin{figure*}
    \centering\includegraphics[width=1\textwidth]{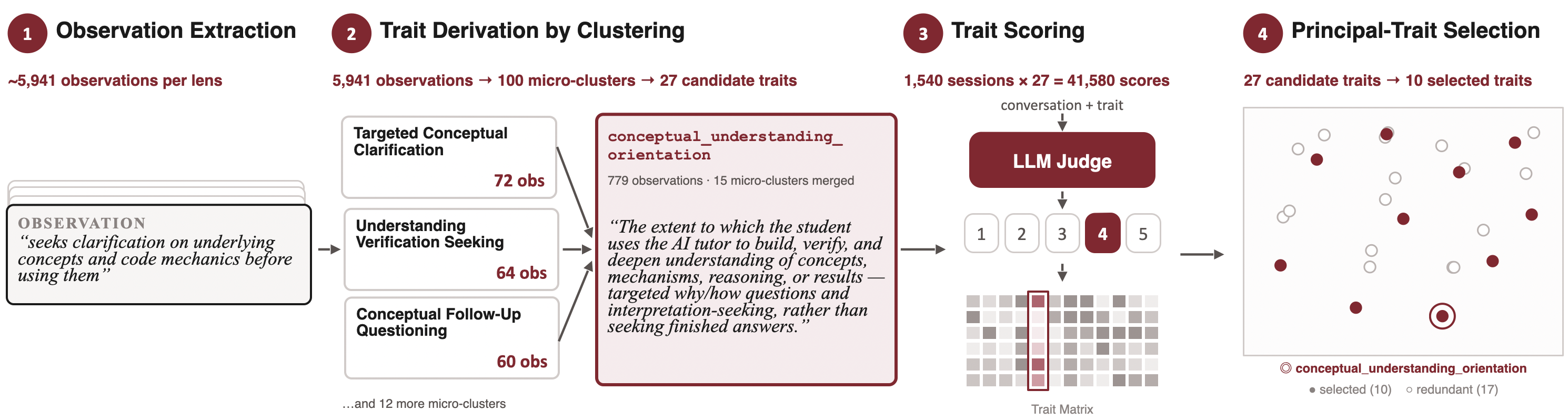}
\caption{\textbf{PTA visualized stage-by-stage}. We first derive a large set of observations from human-AI collaboration sessions, which are iteratively refined into a smaller set of candidate traits. We then score each session against these traits and select a smaller set of explanatory and distinct principal traits.}    \label{fig:pta-pipeline}
\end{figure*}

In this work, we propose a data-driven method to derive informative \textit{traits} in human-AI collaborative coding sessions and use this method to analyze two real-world conversation datasets. Our approach, \textbf{Principal Trait Analysis (PTA)}, is a principal component analysis (PCA)-inspired method that extracts behavior observations from raw conversations, derives interpretable trait dimensions by clustering, scores each human collaborator on every trait, and selects the most distinguishing traits. We extract these traits through a series of unsupervised stages, via LLM-based prompting and text-embedding-based clustering steps, to identify traits that explain common collaborator behavior patterns. This process intuitively mirrors how PCA derives principal component vectors that explain the most amount of variance in datasets. We then perform an outcome-aligned evaluation that measures whether the derived traits explain and predict downstream outcomes. We conduct experiments on two datasets, including a student-AI tutor dialogue dataset and a professional coder-AI coding agent interaction session trace corpus, showing both shared and setting-specific structure in effective LLM use. We find that our approach yield traits that are often significant in explaining outcomes. However, we caution that these traits may not fit the definition of skills, due to limited generalizability across collaborative settings and inconclusive trends in how they evolve over time. 

\section{Methodology} \label{sec:method}

\begin{figure*}[t]
\centering
\includegraphics[width=1\textwidth]{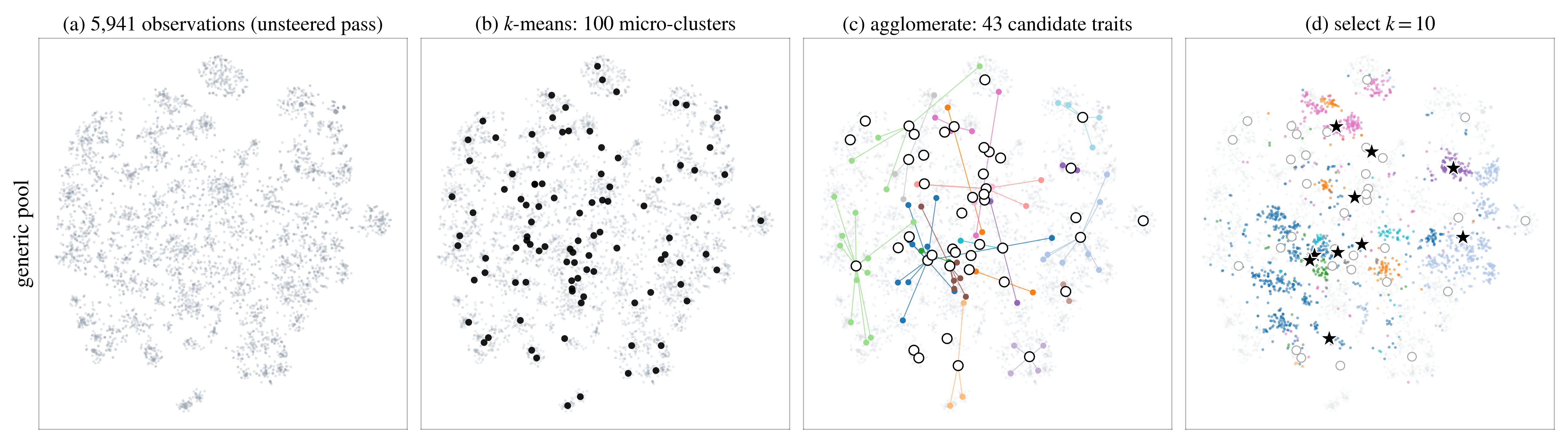}
\caption{\textbf{PTA clustering stages visualized}. Panels show one shared t-SNE projection of all interaction text embeddings from the StudyChat pool. Raw observations (a) are clustered into $k=100$ micro-clusters (b), $\bullet$ indicates cluster centroids. Micro-clusters are agglomeratively merged to remove duplicates and form candidate traits (c), $\circ$ indicates candidate centroids and lines indicate which micro-clusters are merged. Finally the selection algorithm chooses $k{=}10$ principal traits (d), $\star$ indicates selected traits.}
\label{fig:derivation-studychat}
\end{figure*} 

PTA, a Principal Component Analysis (PCA)-inspired method, produces a representative set of textual descriptions we call \emph{principal traits} from human-AI collaborative conversation sessions. Each trait details a behavior pattern present in the corpus, analogous to principal component vectors derived from PCA. Moreover, PTA also produces a "trait score" for each session that enables session-level and human-level behavioral analysis. In Figure~\ref{fig:pta-pipeline}, we visualize the 4 stages of PTA: i) behavior observation extraction, ii) observation clustering, iii) trait scoring, and iv) principal trait extraction. In this section, we detail our methodology.



\paragraph{Notation.}
We first define notation shared across different stages of PTA. We consider $M$ total collaborative sessions $\mathcal{S} = \{s_1, \ldots, s_M\}$, each authored by a human user $u(s_i) \in \mathcal{U}$, where $s_i$ denotes the textual interaction trace of a human-AI session. From $\mathcal{S}$, PTA produces a set of $K$ principal traits $C^*=\{c^*_1 \ldots, c^*_K \}$ and a trait-score matrix $\mathbf{Z} \in \{1, \ldots, 5\}^{M \times K}$, where $z_{ij}$ represents to what extent session $s_i$ exhibits principal trait $c^*_j$. In addition, we use a text-embedding model $E(\cdot) \in \mathbb{R}^d$ to embed text in an LLM $L(\cdot)$'s embedding space to calculate textual similarity, for clustering and ranking purposes.

\paragraph{Stage 1: Behavior observations.}
In this stage, our goal is to propose a set of surface-level behavior \emph{observations} for human-AI interaction sessions. We prompt an LLM to read the session and emit a small set of brief behavior observations (up to 5) corresponding to the human's behavior in this session, i.e., $L(s_i) \rightarrow \{o_1, \dots o_{K_{0}}\}$, $K_0 \leq 5$. We experiment with two prompting approaches, \textit{generic} without any specific guidelines and \textit{lens-based} observations, where the LLM is provided a textual description of a relevant theory towards understanding behavior in a particular context, e.g., self-regulation theory in student-AI tutoring sessions. Full prompts for observation extraction according to all lenses appear in Appendix~\ref{app:prompts}.

\paragraph{Stage 2: Trait derivation by clustering.}
In this stage, we form a set of candidate traits $\mathcal{C}=\{c_1 \dots c_K\}$, by aggregating all observations across all sessions. These candidate traits are textual representations of commonly observed behaviors that contain a name and description; Table~\ref{tab:table2_all_datasets_covariance} (Appendix \ref{app:significant-traits}) shows examples. To form these candidate traits, we perform agglomerative clustering and a text de-duplication step to form descriptive patterns of common behavior. First, we perform top-down clustering on each lens observation set by embedding every observation with a sentence encoder $E(o_i)=\mathbf{e}_i$ and clustering the embeddings via $k$-means to form $K_1$ micro-clusters \cite{ahmed2020kmeans}. We then use an LLM to name each cluster by passing $K_2$ observations whose embeddings are closest to the cluster centroid: we prompt the LLM to name the cluster and provide a short description of summarizing the behavior, i.e., $L(\mathbf{o}'_i)\rightarrow \omega_i$, where $\omega_i$ is a textual description of a cluster based off cluster centroid nearest-neighbors $\mathbf{o}'_i$. Second, we perform bottom-up agglomerative clustering on the micro-cluster \textit{description} embeddings $E(\omega_i)=\boldsymbol{\omega}_i$ to combine similar clusters, until we reach a threshold of $K_3$ candidate traits. Finally, we merge any near-duplicate candidates, whose definition embeddings exceed a cosine similarity threshold $\tau$. This clustering and reduction process is repeated for all lens and all candidates, which we then combined into a set of $K_4$ candidate traits $\mathcal{C}$. Figure~\ref{fig:derivation-studychat} (b) and (c) shows the resulting reduction: thousands of raw observations reduced to hundreds of micro-clusters and then further into dozens of named candidate traits. By combining these steps, we create a set of diverse candidate traits that reflect common human behavior across sessions. 

\paragraph{Stage 3: Trait scoring.}
In this stage, we from a trait-score matrix $\mathbf{Z} \in \{1 \dots 5\}^{M \times K_4}$ from the candidate traits $\mathcal{C}$, where $z_{ij}$ represents the extent to which a human exhibits behavioral trait $j$ in session $i$. 
To compute these scores, we use an LLM-as-a-judge approach \cite{zheng2023judge} for every (session, trait) pair, to rate how the human exhibits the trait on a 5-point Likert scale, $L(s_i, c_j) \rightarrow z_{ij} \in \{1,\ldots,5\}$. We rate each trait in a separate LLM call to prevent judge bias towards a particular polarity within a session. We also average $\mathbf{Z}$ over sessions \textit{per-human} and form another score matrix $\mathbf{Z'} \in \mathbb{R}^{|\mathcal{U}| \times K}$, where $z'_{ij}$ is the mean score score on trait $j$ across all sessions involving human $u_i$. We later use this score matrix for per-user behavioral analysis. Intuitively, this scoring process is analogous to computing the loadings for each principal component in PCA, where each LLM call is a textual analog of performing a projection of a data sample onto the direction of a principal component. We detail the scoring prompts used in this stage in Appendix~\ref{app:scoring-prompts}.

\paragraph{Stage 4: Principal trait selection.}
In this stage, we select a subset of \(K\) principal traits \(C^* = \{c_1^*, \ldots, c_K^*\}\) from the candidate traits \(C\), using the human-level score matrix \(Z'\). Inspired by PCA, we aim to select traits that exhibit high variance across humans while penalizing traits whose scores co-vary with already selected traits. Unlike PCA, which iteratively constructs orthonormal principal components, our approach selects and ranks existing textual traits built above one-by-one, based on their numerical scores across users.

For each candidate trait \(c_j \in C\), we compute its sample variance \(v_j = \operatorname{Var}(z'_j)\), where \(z'_j\) contains the scores for trait \(c_j\) across humans. Higher variance indicates greater differences in how humans exhibit the trait. Starting from \(C^*_0 = \emptyset\), we greedily select, for \(k = 1, \ldots, K\),
\[
c_{k}^{*} = \operatorname*{arg\,max}_{c_j \in C \setminus C^{*}_{k-1}}
\frac{\operatorname{Var}(z'_j)}{1 + \frac{1}{k-1}\sum_{\ell=1}^{k-1} \left|\operatorname{Cov}(z'_j,\, z'_{c_\ell^{*}})\right|},
\]
where \(C^*_{k-1} = \{c_1^*, \ldots, c_{k-1}^*\}\) and the empty sum for \(k=1\). We first select the candidate with the highest variance. We then greedily select the next trait: we divide each remaining candidate's variance by one plus its mean absolute covariance with the previously selected traits, and then select the one that maximizes this value. This design penalizes traits whose scores co-vary with those already selected (analogous to orthogonality between principal components in PCA), while favoring traits that separate humans. We repeat this process until \(K\) traits are selected.

\section{Experimental Setup}
\label{sec:setup}

In this section, we detail our experiments to evaluate the effectiveness of PTA in deriving traits in two different human-AI collaboration settings.

\subsection{Datasets}
The first dataset we analyze is \textbf{StudyChat}~\cite{studychat}, a collection of conversations between students and an LLM tutor in a university programming-based artificial intelligence course. We analyze 1{,}540 tutoring sessions authored by 171 students across two semesters. Students complete seven programming assignments and three examinations over a semester. 
The two semesters involve different students, assignments, and exams, so we analyze outcomes on each semester separately but derive principal traits from the entire dataset.

The second dataset we analyze is \textbf{SWE-Chat}~\cite{swechat}, a collection of conversations between professional developers and AI-powered coding agents working on real software repositories. We analyze 2{,}774 sessions that come from developers with at least two sessions. Each session has a success score that indicates how well the human-AI collaboration completed the task. In SWE-Chat, each session addresses a distinct software task rather than a shared assignment across all students in StudyChat, which we account for in our evaluation metrics.


\subsection{Metrics}
We measure whether the derived principal traits explain collaborator outcomes beyond a standard prior of average outcome on previous scored tasks~\cite{hallberg2018pretest}. We define our outcome target variable and features to account for the different settings across the two datasets below.



\paragraph{StudyChat}
In StudyChat, students complete a shared set of assignments and examinations, so we use exam scores, which reflects student learning outcomes, as our prediction target. We examine two sets of features: i) using only assignment scores prior to each exam as features, and ii) augment these features with averaged trait scores in all sessions recorded before each exam. We treat each of the two later examinations, $e_2$ and $e_3$, as a separate target; therefore, each of the $171$ students contributes two samples, one per exam, for $342$ target outcomes in total. Since the effective sample size is too small to support a held-out partition, we therefore report explanatory improvement $R^2$ from in-sample fits.

\paragraph{SWE-Chat: Predictive Difficulty-Aligned Regression}
In SWE-Chat, each session addresses a distinct task, ranging from quick webpage fixes to advanced distributed backend deployments, so we normalize outcomes by task difficulty and compute features from \textit{per-session} trait scores. Specifically, we fit an item response theory (IRT) \cite{hambleton2013item} type model, $P(\text{success}) = \sigma\big(\mathbf{w}_a^\top \mathbf{u} - \mathbf{w}_d^\top \mathbf{t}\big)$, in which collaborator behavior traits $\mathbf{u}$ are related to ability and task-property traits $\mathbf{t}$ are related to task difficulty~\cite{deboeck2004eirm}. We simply minimize the binary cross entropy loss between the continuous session success score (normalized to $[0,1]$) and this predicted success probability. 
Since each session carries its own outcome, this setup yields one regression target per session, and the sample is large enough that we report predictive $R^2$ on held-out sessions. We perform cross validation to create 4 sets of train-test splits of our data, divided across human users. We use the train split to train the IRT outcome model and report the average held-out $R^2$ improvement across all folds.

\subsection{Experimental Details}
\paragraph{Extraction Lenses.}
We extract observations under a set of lenses, each of which frames the extraction prompt around a body of domain-specific theory. For StudyChat, we use four lenses drawn from education research: self-regulated learning~\cite{zimmerman} combined with the ICAP engagement framework~\cite{chiwylie}, the 4D AI-fluency competencies~\cite{dakan4d}, question sophistication~\cite{hackl2026}, and academic help-seeking~\cite{aleven2016}. We adapt the third lens from the technical dimension of the AI Literacy Heptagon, treating the precision and depth of a student's questions as a conversational proxy for that dimension~\cite{hackl2026}. 
For SWE-Chat, we use four lenses drawn from software-engineering practice: specification quality, verification and oversight, agency and control, and craftsmanship and process discipline. We frame the first, second, and fourth around knowledge areas of the Software Engineering Body of Knowledge~\cite{swebok}, covering requirements, testing, and construction and quality. We frame the third around a taxonomy of human-AI collaboration in software engineering~\cite{treude2025} and a study of how professional developers supervise coding agents rather than delegate to them wholesale~\cite{huang2025control}.



\paragraph{PTA Hyperparameters.} We detail the hyperparameters used in our experiments. We set $K=10$, $K_1=100$, $K_2=10$, and $\tau=0.85$ for both datasets. We set $K_3$ the aggregation target to $45$ and $12$ for StudyChat and SWE-Chat respectively to account for the differences in the sizes of the datasets.

\paragraph{Language Models.} 
For observation extraction and trait scoring we use the GPT5.5 to produce observations, cluster names, and de-duplication~\cite{gpt55}, and deploy a self-hosted \texttt{Qwen3-4B} model~\cite{qwen3} for scoring. For text-embeddings we use ModernBERT~\cite{modernbert} to form mini-clusters from the observations. 

\paragraph{Baselines.}
We compare PTA derived traits against two feature sets defined for StudyChat. \textbf{Dialogue acts} uses the counts of the 8 broad dialogue-act categories per turn, provided in the dataset which we derived from manual coding and observations of the sessions. \textbf{Bloom's taxonomy} uses the 6 levels of the revised Bloom's taxonomy~\cite{krathwohl2002} as an alternative set of traits. We prompt our scoring model to measure the degree to which each of the levels of the schema was present in the dataset through the same scoring prompt structure as the candidate traits and use the scores as regression features. See Appendix~\ref{app:blooms-baseline} for further details. For SWE-Chat, we could not find an established taxonomy of predictive dialogue traits similar to what the learning sciences provide for software engineering task outcomes. Therefore, the baseline is simply the average success score across all sessions in the training set for each human user. 
\subsection{Human Evaluation}
To assess whether PTA-derived traits can be consistently identified by human evaluators and how closely their ratings align with LLM-generated scores, we conduct a human evaluation on a subset of StudyChat. 
\textbf{Annotation Procedure.} We select 10 StudyChat sessions and recruit two volunteer evaluators experienced in computer science tutoring. Each evaluator independently rates all sessions on the same 10 PTA-derived principal traits from the generic pool, using a five-point Likert scale. Each rating indicates the extent to which the student's behavior in a session exhibits the corresponding trait. We provide evaluators with the same scoring guidelines used by the LLM judge (Appendix \ref{app:scoring-prompts}). In total, we collect 200 human ratings across 100 session–trait pairs, each with a corresponding LLM-generated score.
\textbf{Metrics.} We measure agreement between human evaluators and between human and LLM-generated ratings using linearly weighted Cohen's kappa \cite{cohen1968weighted}, exact agreement, within-one-point agreement, mean absolute error (MAE), and Spearman correlation \cite{spearman1904proof}. We also report the mean difference between LLM and human ratings to measure scoring bias.

\section{Results and Analysis}

\input{tables/all_dataset_covariance}

\label{sec:results}

We evaluate whether PTA-derived traits explain or predict collaboration outcomes, how their scores change over time, and how consistently they can be rated by human evaluators and LLMs. 

\subsection{PTA Traits Validity}

Table~\ref{tab:incremental_validity_new} shows how much PTA-derived traits improve the explanatory power of regression models beyond the baseline features. On StudyChat, adding 10 traits from the generic pool significantly increases \(R^2\) in Fall 2024 (\(\Delta R^2 = 0.136\), \(p < 0.01\)), but not in Spring 2025 (\(\Delta R^2 = 0.055\), \(p = 0.144\)). Traits from the lens-ensemble pool, as well as the dialogue-act and Bloom's taxonomy baselines, do not yield statistically significant improvements in either semester. These results show that PTA-derived traits explain additional variation in exam outcomes in Fall 2024, but the improvement is not statistically significant in Spring 2025. Therefore, we cannot yet conclude that these traits generalize across course settings. 
%
We also see that learned principal traits outperform coded dialogue act or Bloom's taxonomy baselines. This observation suggests that baseline coding schemes are either too specific to explain overall behavior in the case of dialogue acts and too general in the case of Bloom's. Instead, our traits strike a better balance. 


On SWE-Chat, PTA-derived traits significantly improve the prediction of session outcomes beyond the baseline using developers' prior success. As shown in Table 1, traits from the generic and lens-ensemble pools increase \(R^2\) by 0.034 and 0.014, respectively (both \(p < 0.01\)). These results suggest that PTA-derived traits capture behavioral information relevant to coding session outcomes beyond prior performance. However, the overall \(R^2\) values are lower than those on StudyChat, although the different outcome variables and evaluation procedures make direct comparisons difficult. While these results demonstrate the applicability of PTA across different collaboration settings, the lack of established behavioral-trait baselines for SWE-Chat limits our ability to assess its relative effectiveness.

\subsection{Significant Traits}
In Table~\ref{tab:table2_all_datasets_covariance} in Appendix~\ref{app:significant-traits}, we report the individually significant traits derived from the regression analyses in the previous section. Since our traits are text-based, we see that these traits are not only explanatory, but are also interpretable. 


On StudyChat, we observe that behaviors that encourage conceptual understanding are positively correlated with exam outcomes, whereas goal-directed and task-delegating behaviors are negatively correlated with outcomes. Specifically, we see that \texttt{question\_granularity}, \texttt{question\_context\_elaboration}, and \texttt{help\_request\_context} are traits positively correlated with exam outcomes and indicate deep conceptual engagement. However, \texttt{task\_context\_specificity} is negatively correlated with outcomes, which appears to contradict the results on other traits. One possible explanation is that certain types of help-seeking behaviors are negatively correlated with outcomes, and these questioning behaviors may be confounded by student ability. Notably, these unsupervised findings corroborate with manually-coded observations on the StudyChats dataset that conceptual questioning dialogue acts were positively correlated with outcomes~\cite{studychat}. The remaining significant traits are all negatively correlated with outcomes and may suggest a pattern of task-delegation behaviors, where the student asks the LLM to complete a portion of the assignment. This result suggests that these traits emerge when students are not engaging with learning outcomes, which is aligned with prior work analyzing this dataset~\cite{rinja2026vibecoding}. 

On SWE-Chat, we observe that behaviors where developers provide a clear role for the LLM are positively correlated with outcomes, whereas evidence-based workflows are negatively correlated with outcomes. 
Traits where the user gives specific, scoped tasks to the agent result in positive outcomes such as \texttt{workflow\_control\_delegation}, and \texttt{contextual\_workflow\_guidance}. However, when the developer provides loosely-defined roles and instead gives constraint-based guidelines or relies on the coding agent's interpretation of evidence, the collaboration often results in negative outcomes, such as \texttt{diagnostic\_evidence\_provision} and \texttt{evidence\_driven\_failure\_validation}. 


\subsection{Generalizability and Temporal Analysis}

For a trait to qualify as a valid skill, it needs to be generalizable across settings. We therefore transfer regression models between the Fall 2024 and Spring 2025 StudyChat cohorts without refitting (Appendix~\ref{app:cross-semester}). Three of the four transfer conditions yield negative changes in \(R^2\) relative to the prior-performance baseline, while the remaining condition yields a small positive change. These results suggest that the relationships between PTA-derived traits and exam outcomes do not consistently generalize, even for different cohorts of the same course.

Another important characteristic of skills is that they improve with practice \cite{cen2006lfa}. We examine whether PTA-derived traits lead to plausible learning curves by analyzing how their scores change over time. Figure \ref{fig:over_time} (Appendix \ref{app:temporal-analysis}) shows the average scores of traits significantly associated with outcomes, across assignments for StudyChat and successive sessions for SWE-Chat. On StudyChat, most traits remain relatively stable, although \texttt{question\_granularity} and \texttt{question\_context\_elaboration} show modest increases over the semester. Figure \ref{fig:studychat-traitplots} (Appendix \ref{app:temporal-analysis}) further shows broadly similar trajectories across Fall 2024 and Spring 2025, suggesting that these behavioral patterns are not unique to one cohort. 
For SWE-Chat, we observe a stable or slightly declining trend over time for all traits. This stability suggests that behavior in subsequent sessions for software developers does not deviate much from prior sessions: there do not seem to be any clear indications of behavior change or developers learning skills. 
We attribute these findings to two possible causes: First, no feedback on AI usage is provided to the user in both datasets to help them develop human-AI collaboration skills. Second, since humans interact with AI frequently, these datasets that are captured only across a short time period may coincide with a flat region of the learning curve. The second explanation is more likely for SWE-Chat, since for professional developers, their skills are likely to be more stable than students. 

Overall, these generalizability and temporal investigation results show that, while PTA can discover human-AI collaboration behavior traits that are explanatory of user behavior and predictive of collaboration outcomes and future success, they do not fit the traditional definition of skills in the literature \cite{koedinger2016testing}. 


\subsection{Human Evaluation}\label{sub_sec:human_eval}
\begin{table}[t]
  \centering
  \scriptsize
  \setlength{\tabcolsep}{2.5pt}
  \scalebox{1.15}{
  \begin{tabular}{@{}lcccccc@{}}
    \toprule
    \textbf{Comparison} & \textbf{Exact} & \textbf{$\leq 1$} & \textbf{$\kappa_w$} & \textbf{MAE} & \textbf{Bias} & \textbf{$\rho$ }\\
    \midrule
    Human--human & 43\% & 70\% & 0.447 & 0.94 & --- & 0.621 \\
    Human--LLM (Rater 1) & 26\% & 60\% & 0.225 & 1.30 & +0.56 &0.348 \\
    Human--LLM (Rater 2) & 33\% & 75\% & 0.340 & 1.00 & +0.76 & 0.598 \\
    \bottomrule
  \end{tabular}}
    \caption{Human evaluation of trait scoring. LLMs and human evaluator themselves have different score scales but agree more on the relative ordering of session-trait pairs.}
  \label{tab:human_judge_agreement_compact}
\end{table}

Table~\ref{tab:human_judge_agreement_compact} shows the agreement between human evaluators and between human and LLM-generated trait scores. Human evaluators show moderate agreement while the human-LLM agreement is lower on weighted $\kappa$. However, Spearman correlation $\rho$ values are generally higher, with good agreement in some cases. This observation suggests that while LLMs and individual human raters have different score scales, they agree more on the relative score ordering across users and sessions. 

Agreement also varies across individual traits, as shown in Table ~\ref{tab:human_trait_agreement} in the Appendix. Traits such as \texttt{conceptual\_clarification\_seeking} and \texttt{error\_evidence\_specificity} exhibit relatively high human agreement, whereas \texttt{task\_context\_specificity} shows minimal agreement. 
Overall, these results provide preliminary evidence that, while the identified PTA-derived traits can be meaningful to humans, only some can be reliably scored by LLMs. 
We emphasize the need for further human evaluation studies given the small sample size of our analysis.

\section{Related Work}

\subsection{LLM Clustering and Topic Modeling}
LLMs have been widely applied to derive interpretable structure from text corpora, and have been evaluated directly against classical topic models such as LDA~\cite{mu2024}. Prompt-based topic modeling pipelines generate and refine natural-language topic labels~\cite{topicgpt, bcllm}, and goal-conditioned variants discover structure relevant to a stated analysis purpose~\cite{goalex, instructlf}. Closer to our derivation stages, prior work uses LLM feedback to resolve ambiguous cluster assignments and choose granularity~\cite{clusterllm}, or inductively generates labels before hierarchically clustering them into broader themes~\cite{hicode}. Recent work had also compared LLM based inductive coding against human evaluations\cite{parfenova-etal-2025-emergent} and proposed methods for aggregating and evaluating open code using LLM-guided clustering \cite{chen-etal-2026-computational}. Related work also proposes methods of visualizing these derived structures in dashboards for monitoring and manual analysis of text corpora~\cite{clio,lloom}. Large-scale chat datasets have attracted growing interest as LLM collaboration expands across tasks and conversation datasets become available.
~\cite{swechat,wildchat, shao2026collabskill, studychat}. Therefore similar analysis pipelines are now being applied to mine insights from chat datasets~\cite{tntllm,clio}. Analyses of AI-assisted IDE conversations identify recurring patterns of progressive specification, delegation, and active control of coding agents \cite{tang2026codingagentsfailusers}. PTA's trait-derivation stages build on this machinery, using LLMs to derive observations from chat datasets, conditioned on the collaborative context and optionally a theory lens.

\subsection{LLM Features for Outcome Prediction}
LLM-scored features are increasingly used for outcome prediction, building on embedding-based approaches \cite{althoff2016}. Applications include predicting diagnoses from clinical notes\cite{chill,bcllm}, intervention outcomes from patient reports~\cite{d5}, ratings from business reviews~\cite{tbm}, and fMRI responses from questionnaire items~\cite{qaemb}. Specific to conversation datasets, prior work has used LLM-derived features to score and predict user satisfaction~\cite{spur}. However, while LLM-based feature extraction and scoring is convenient, it is unclear which features are more informative than others without directly regressing on task outcome. In our work, we explore a methodology that extracts traits as downstream predictive features, scores conversations based on these traits, and determines the most informative traits prior to outcome regression. Trait derivation thus remains independent of outcome labels, while the resulting traits can explain variance in downstream outcomes.


\subsection{Education Tutoring Dialogues}
The relationship between conversation behavior and learning outcomes has a long history in education research. Prior work linked manually coded tutoring dialogue acts to learning gains\cite{litman2006,graesser2016}, modeled help-seeking in tutoring systems~\cite{aleven2016}, and discovered latent dialogue modes ~\cite{boyer2011}. More recently, LLMs have been explored for capabilities in education tasks such as automated scoring and content generation. In this line of work, related studies have scored dialogues on predefined psychological constructs such as persistence~\cite{ober2026} and mined student--AI chat logs for knowledge gaps~\cite{mininggold}. 

In a parallel thread, educators are now interested in teaching effective usage of AI tools, resulting in a growing number of ``AI literacy'' frameworks which establish effective principles for AI usage with learning and professional objectives~\cite{longmagerko}. Many of these frameworks are grounded in pre-existing learning and tutoring frameworks such as self-regulated learning~\cite{zimmerman} and the Interactive, Constructive, Active, and Passive (ICAP) framework for cognitive engagement~\cite{chiwylie}. However, since communication styles between AI and human tutors are distinct, new frameworks are being developed to characterize these styles and related competencies~\cite{dakan4d}. Assessing these competencies is itself an open problem: recent work finds substantial misalignment between self-reported and objective-based measures of AI literacy~\cite{zhang2026ailit}. However, much of this work relies on manual coding and inspection of conversation data which does not scale readily, especially given the rapid development of LLM systems.

\section{Conclusions and Future Work}
In this paper, we proposed Principal Trait Analysis (PTA), a method for deriving behavioral traits from human-AI collaboration dialogues. We evaluated PTA in tutoring and software development settings, finding that the derived traits help explain variation across users but do not consistently generalize or evolve over time, falling short of the definition of skills. 
%
Avenues for future work include evaluating PTA across more human-AI collaboration settings and larger datasets to assess the generalizability of the derived traits. We can also investigate whether providing feedback to human users on some traits lead to improvement over time. 

\section*{Limitations}
Our work has several limitations. First, although we evaluate Principal Trait Analysis (PTA) on two human-AI collaboration datasets, both are situated within programming-related contexts. PTA has yet to be evaluated in other collaborative settings. However, public human--AI dialogue datasets that are linked to downstream outcomes (exam scores, task success) remain rare, and StudyChat and SWE-Chat are among the few that provide both. Moreover, these datasets still represent distinct setting with different populations, tasks, and outcome measures. Our results demonstrate the applicability across these two settings. 

Second, PTA relies on LLM-generated observations and trait scores, which may be sensitive to model choice, prompting strategies, and clustering hyperparameters. We mitigate these concerns in several ways. We mitigate these concerns by evaluating both generic and theory-guided observation extraction and assessing scoring reliability through human evaluation. We also use a small open-weight model with temperature 0 to improve reproducibility. Importantly, trait derivation and selection are performed without access to outcome gains, reducing the risk of outcome-driven tuning.

Finally, our temporal analyses are limited to the observation periods available in the two datasets. The absence of consistent improvements in trait scores does not necessarily imply that these behaviors cannot be learned or developed. Longer-term studies, particularly those involving explicit feedback or interventions, are needed to determine whether the identified traits represent learnable skills and whether improving them leads to better collaboration outcomes.

\section*{Ethical Considerations}
Our work identifies behavioral traits from human-AI collaboration traces which may reflect differences in user expertise or context. These traits should therefore not be used for high-stakes assessments, such as grading or employee evaluation, without further validation.

\bibliography{custom}
\clearpage
\appendix
\section{Significant Traits with Definitions}
\label{app:significant-traits}
In Table~\ref{tab:table2_all_datasets_covariance}, we report the individually significant traits derived from the regression analyses. Since our traits are text-based, we see that these traits are not only explanatory, but are also interpretable.

\begin{table*}[!t]
\centering
\small
\setlength{\tabcolsep}{3pt}
\begin{tabularx}{\textwidth}{@{}>{\raggedright\arraybackslash}p{3.8cm} >{\raggedright\arraybackslash}X r@{}}
\toprule
Trait & Description & $\beta$ ($p$) \\
\midrule
\multicolumn{3}{l}{\textbf{StudyChat---lens-ensemble pool}} \\
\texttt{debugging\_\allowbreak context\_\allowbreak specificity} &
How concretely the student frames troubleshooting requests, including exact error messages, observed outputs, relevant code or artifacts, surrounding context, and attempted fixes &
-0.434 (.013)$^{\mathrm{F}}$ \\
\addlinespace
\multicolumn{3}{l}{\textbf{StudyChat---generic pool}} \\
\texttt{question\_\allowbreak granularity} &
The student frames help requests as narrow, specific questions about particular code elements, syntax, APIs, or implementation choices versus broader, more contextual task descriptions &
+0.376 (.012)$^{\mathrm{F}}$ \\
\texttt{help\_\allowbreak request\_\allowbreak context} &
How much relevant context a student provides when asking for help &
+0.338 (.029)$^{\mathrm{F}}$ \\
\texttt{question\_\allowbreak context\_\allowbreak elaboration} &
A student provides context, goals, prior attempts, and details when asking the AI tutor for help &
+0.337 (.020)$^{\mathrm{F}}$ \\
\texttt{task\_\allowbreak context\_\allowbreak specificity} &
The student provides concrete, task-specific materials or details---such as code, data, schemas, drafts, or outputs---for the tutor to work from &
-0.379 (.036)$^{\mathrm{F}}$ \\
\addlinespace
\multicolumn{3}{l}{\textbf{SWE-Chat---lens-ensemble pool}} \\
\texttt{constraint\_\allowbreak scope\_\allowbreak control} &
The user explicitly bounds the agent's autonomy by specifying scope limits, implementation requirements, file targets, workflow rules, options, or acceptance criteria &
+0.016 (.005) \\
\texttt{verification\_\allowbreak gatekeeping} &
The user explicitly defines, monitors, and enforces tests, checks, acceptance criteria, repository-state inspections, or manual validation as gates before progress, commits, PRs, or completion claims &
+0.014 (.024) \\
\texttt{evidence\_\allowbreak driven\_\allowbreak failure\_\allowbreak validation} &
The user grounds debugging and review in concrete failure evidence such as logs, test output, CI results, reproduction steps, and expected-versus-actual behavior &
-0.029 ($<$.001) \\
\addlinespace
\multicolumn{3}{l}{\textbf{SWE-Chat---generic pool}} \\
\texttt{workflow\_\allowbreak control\_\allowbreak delegation} &
The user directs the AI coding agent's working process, ranging from terse high-level autonomous delegation to highly prescriptive, command-driven workflow orchestration &
+0.023 ($<$.001) \\
\texttt{contextual\_\allowbreak workflow\_\allowbreak guidance} &
The user explicitly supplies or directs the agent to gather task context, inspect repository state, use specific paths or commands, and follow verification or documentation expectations &
+0.011 (.024) \\
\texttt{diagnostic\_\allowbreak evidence\_\allowbreak provision} &
The user reports bugs or failures with concrete, actionable evidence such as reproduction steps, expected-versus-actual behavior, logs, stack traces, commands, screenshots, CI artifacts, or failing inputs &
-0.023 ($<$.001) \\
\addlinespace
\bottomrule
\end{tabularx}
\caption{\textbf{Significant Traits with Definitions.} Traits individually significant ($p < .05$) in the models presented in Table~\ref{tab:incremental_validity_new} trait selection. $\beta$ is the standardized coefficient in the regression on the target.}
\label{tab:table2_all_datasets_covariance}
\end{table*}
\FloatBarrier


\section{Bloom's Taxonomy Baseline Dimensions}
\label{app:blooms-baseline}
Table~\ref{tab:blooms_baseline} defines the six dimensions used in the Bloom's-taxonomy baseline. Bloom's taxonomy is a a hierarchical model used by educators to classify learning objectives, instructions, and assessments by their cognitive complexity ~\cite{krathwohl2002}.

\begin{table*}[!t]
\centering
\small
\setlength{\tabcolsep}{6pt}
\renewcommand{\arraystretch}{1.15}

\begin{tabularx}{\textwidth}{
    @{}
    l
    >{\ttfamily}l
    >{\raggedright\arraybackslash}X
    @{}
}
\toprule
\textbf{Level} &
\textbf{\normalfont Dimension} &
\textbf{Definition} \\
\midrule

Remember &
blooms\_remember &
Student asks for definitions, facts, syntax, or terminology
to be recalled or restated --- e.g., ``what is the function
for X'', ``what does this term mean'' --- without asking
for explanation of underlying mechanism. \\

Understand &
blooms\_understand &
Student asks for explanations, interpretations, summaries,
or restatements in their own terms --- e.g., ``why does this
work'', ``explain what this code does'', ``what does this
output mean''. \\

Apply &
blooms\_apply &
Student uses or requests use of a known procedure, formula,
or method on a concrete problem --- e.g., ``how do I use
this function on my data'', ``apply this method to my case'',
``write code that does X with my variables''. \\

Analyze &
blooms\_analyze &
Student breaks material into parts, compares alternatives,
traces causes, or differentiates components --- e.g.,
debugging by isolating causes, comparing two approaches,
asking how parts of a system relate. \\

Evaluate &
blooms\_evaluate &
Student makes or requests judgments against criteria ---
e.g., checking whether their answer/code/interpretation
is correct, critiquing output quality, weighing whether
an approach is appropriate. \\

Create &
blooms\_create &
Student produces or plans original work by combining
elements into a new whole --- e.g., designing an analysis
plan, drafting original written interpretations,
composing new code architecture rather than requesting
boilerplate. \\

\bottomrule
\end{tabularx}

\caption{\textbf{The Bloom's-taxonomy baseline feature set.}
Six hand-authored rubric dimensions, one per cognitive-process
level of the revised taxonomy~\cite{krathwohl2002}, written as
observable student behaviors in AI-tutor conversations.
These dimensions are specified directly rather than derived,
so Stages 1--2 of PTA do not apply; they are scored by the
same model, prompt, and 1--5 scale as the derived traits
(Appendix~\ref{app:scoring-prompts}), which isolates trait derivation
as the difference between the two feature sets.}
\label{tab:blooms_baseline}
\end{table*}
\FloatBarrier

\section{StudyChat: Cross-Semester Generalizability}
\label{app:cross-semester}
We evaluate the cross-semester generalizability of PTA-derived traits using the Fall 2024 and Spring 2025 StudyChat cohorts. Specifically, we examine whether traits selected from one semester improve exam outcome prediction in the other semester, without refitting the regression model.

\paragraph{Experimental setup.} We conduct bidirectional transfer experiments between the Fall 2024 and Spring 2025 semesters, separately for the generic and lens-ensemble candidate trait pools. For each source semester, we select 10 principal traits using collaborator-level scores aggregated across all available sessions. We then fit a regression model to predict standardized examination outcomes using prior examination performance and the selected trait scores. We evaluate the trained model on the target semester without refitting its coefficients or feature standardization. We report the improvement in out-of-sample \(R^2\) over a prior-examination-only baseline trained on the same source semester.

\paragraph{Results} Table \ref{tab:model_transfer} shows the cross-semester transfer results. For the generic trait pool, transferring the regression model from Fall 2024 to Spring 2025 yields a \(\Delta R^2\) of \(-0.171\), while transfer in the opposite direction yields \(+0.038\). The lens-ensemble pool yields negative \(\Delta R^2\) values in both directions (\(-0.114\) and \(-0.032\), respectively). Although the models show positive in-sample improvements, these improvements do not consistently transfer across semesters. These results suggest that the fitted models have limited generalizability across student cohorts.

\begin{table}[t]
\centering

\small
\setlength{\tabcolsep}{2pt}
\begin{tabularx}{\columnwidth}{@{}>{\raggedright\arraybackslash}p{1.85cm} >{\raggedright\arraybackslash}p{2.25cm} >{\centering\arraybackslash}X >{\centering\arraybackslash}X@{}}
\toprule
Trait pool & Direction &
\shortstack{In-sample\\$\Delta R^2$} &
\shortstack{Transfer\\$\Delta R^2$}  \\
\midrule
Generic & Fall $\rightarrow$ Spring
    & $+0.049$ & $-0.171$ \\
Generic & Spring $\rightarrow$ Fall
    & $+0.131$ & $+0.038$ \\
Lens-ensemble & Fall $\rightarrow$ Spring
    & $+0.042$ & $-0.114$ \\
Lens-ensemble & Spring $\rightarrow$ Fall
    & $+0.099$ & $-0.032$  \\
\bottomrule
\end{tabularx}
\caption{In-sample and frozen transfer performance.}
\label{tab:model_transfer}
\label{tab:transfer_performance}
\end{table}

\section{Data Licensing, Privacy, and Consent}
We conduct our experiments using two publicly available datasets: StudyChat \cite{studychat}, containing student interactions with an LLM tutor in a university programming-based AI course, and SWE-Chat \cite{swechat}, containing interactions between professional developers and AI coding agents. Both datasets are publicly released under licenses specified by their respective providers. We refer readers to the original dataset publications and repositories for the specific licensing terms and permitted uses.

\section{Models and Compute}
Observation extraction, cluster naming, and duplicate merging use GPT-5.5 through an
Azure OpenAI deployment (parameter count undisclosed). Trait scoring uses Qwen3-4B (4.0B parameters) served with vLLM, temperature 0, one (session, trait) pair per call. Observation embeddings for clustering use
\texttt{modernbert-embed-base} (149M parameters). Scoring the reported trait pools took about 581k Qwen3-4B calls, roughly 10 GPU-hours on one NVIDIA H100, plus about 28k GPT-5.5 extraction calls, approximately a few hundred US dollars of API spend.

\FloatBarrier
\section{Human Evaluation Details}
\FloatBarrier
\begin{figure*}
    \centering
    \includegraphics[width=1\linewidth]{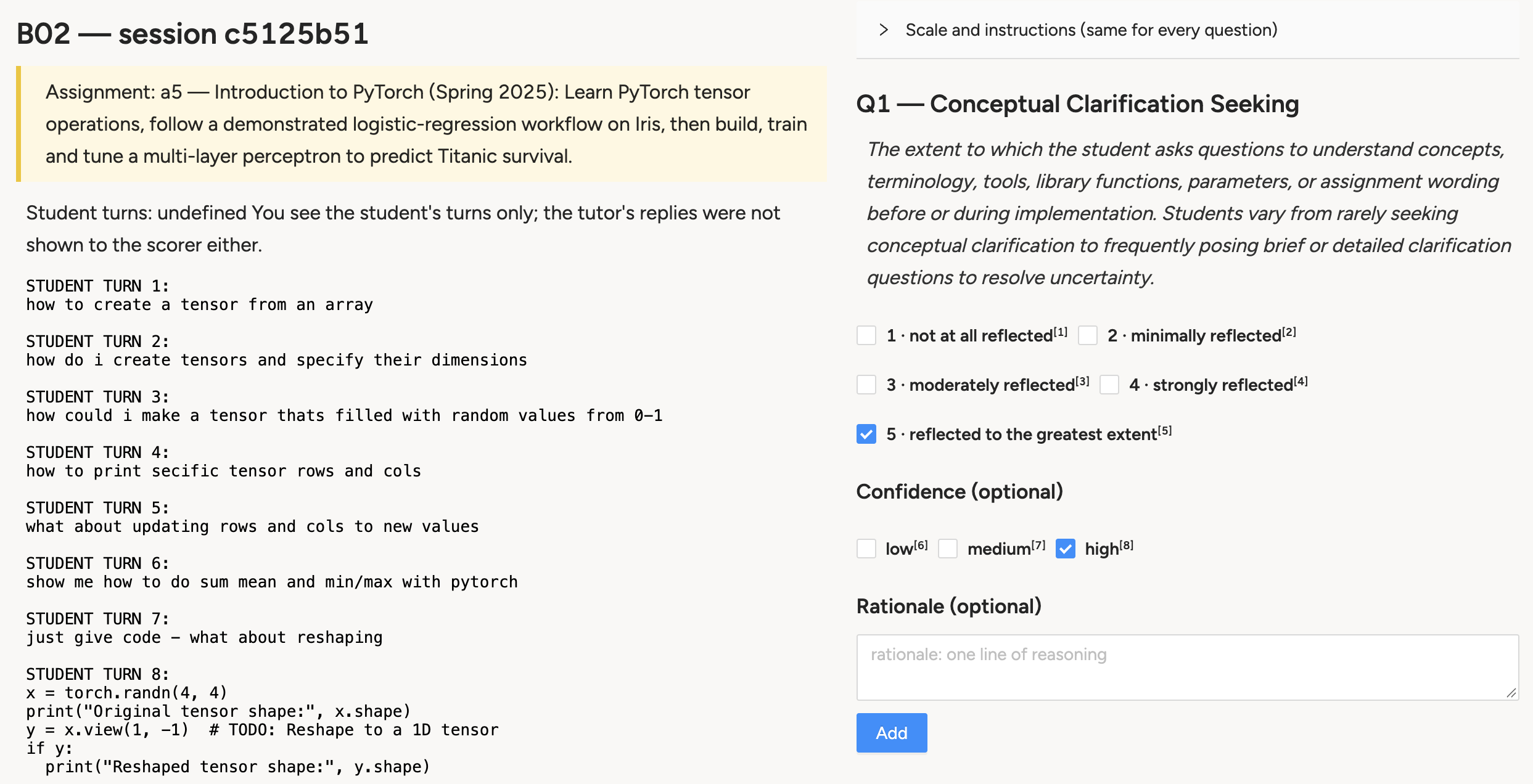}
    \caption{\textbf{Human annotation interface for evaluating PTA-derived traits on StudyChat.} The left panel displays assignment context and student turns, with tutor responses omitted. The right panel presents the trait definition and a five-point Likert scale for rating the extent to which the student’s behavior exhibits the trait, along with optional confidence and rationale fields.}
    \label{fig:interface}
\end{figure*}
\paragraph{Recruitment.} We recruited two volunteer annotators with experience in college level computer science tutoring to evaluate the interpretability and reliability of PTA-derived traits. Both annotators participated voluntarily and received no financial or other compensation for their contributions. We obtained written informed consent from all human evaluators prior to their participation in the study.

\paragraph{Instructions.} We provided a document that explains the context of our research, defines traits, overviews the dataset, and demos the interface as see in Figure \ref{fig:interface}. We then provide the following instructions:
\begin{itemize}
    \item You will be given a session from the StudyChat dataset.
    \item StudyChat has logs of students using ChatGPT for their assignments.
    \item You will be provided the trait definition in each question for references. 
    \item You will be asked to rate each problem on a likert scale from 1-5 for each trait. The definition of each point in the scale is provided. 
    \item We will only provide student turns, so judge based of what the student says. 
\end{itemize}

\paragraph{Annotations.} Each annotator independently evaluated the same 10 StudyChat sessions across 10 principal traits, producing 100 ratings per annotator and 200 ratings in total. Annotators were provided with the trait definitions and the same five-point Likert-scale scoring guidelines used by the LLM judge, as described in Section \ref{sub_sec:human_eval} and Appendix \ref{app:scoring-prompts}. Figure \ref{fig:interface} shows the interface we used for our study. 
\FloatBarrier
\section{Human Evaluation Comprehensive Results Tables}
\label{app:human-evaluation}

Table ~\ref{tab:human_trait_agreement} shows that agreement varies across individual traits. Traits such as \texttt{conceptual\_clarification\_seeking} and \texttt{error\_evidence\_specificity} exhibit relatively high human agreement, whereas \texttt{task\_context\_specificity} shows minimal agreement.

\input{tables/human_evaluation_tables}
\FloatBarrier


\section{Principal Traits Temporal Analysis}
\label{app:temporal-analysis}
\begin{figure*}[!t]
    \centering
    \includegraphics[width=1\linewidth]{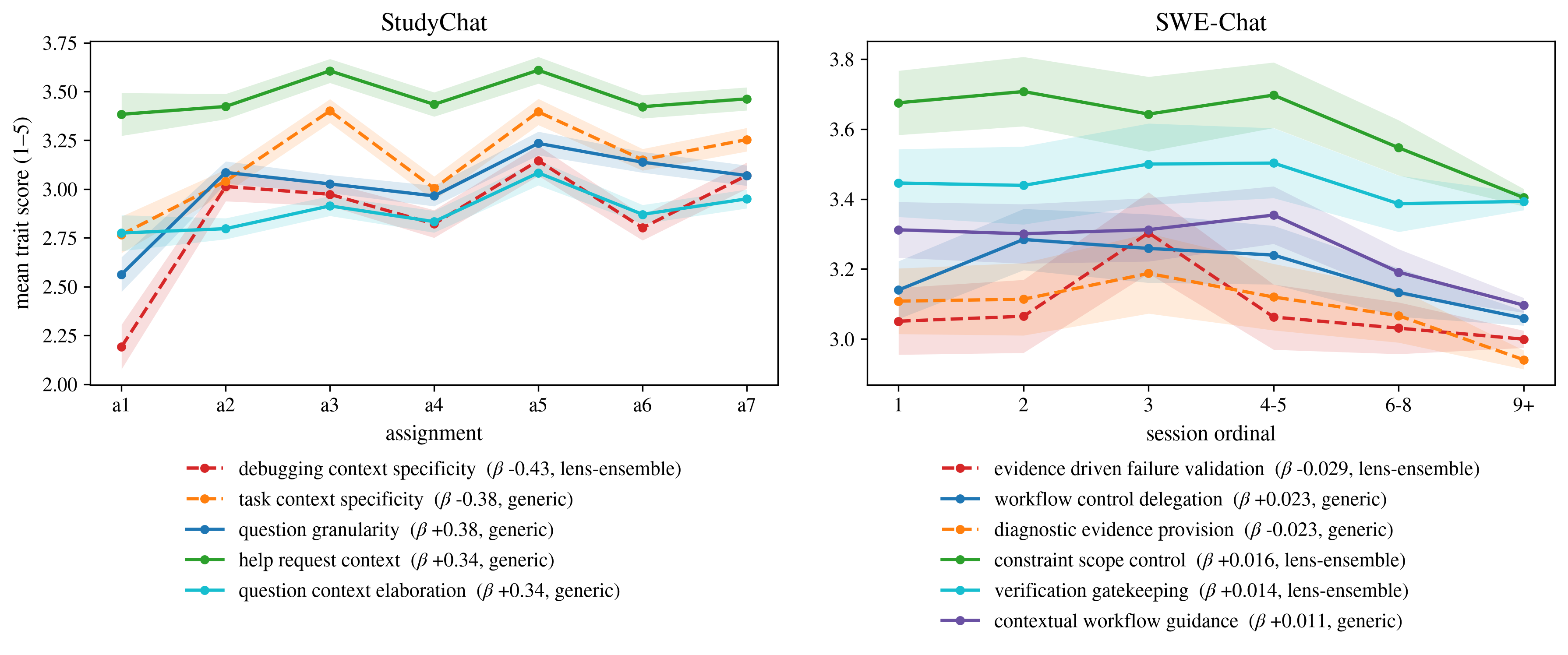}
    \caption{\textbf{Trait scores over time.} Mean session score (1 to 5 scale) of every trait that is individually significant in the Table \ref{tab:incremental_validity_new} model, shown over the course of the semester for StudyChat (left, by assignment, both cohorts pooled) and across successive sessions for SWE-Chat (right, by within-developer session ordinal).}
    \label{fig:over_time}
\end{figure*}

\begin{figure*}[!t]
    \centering
\includegraphics[width=1\linewidth]{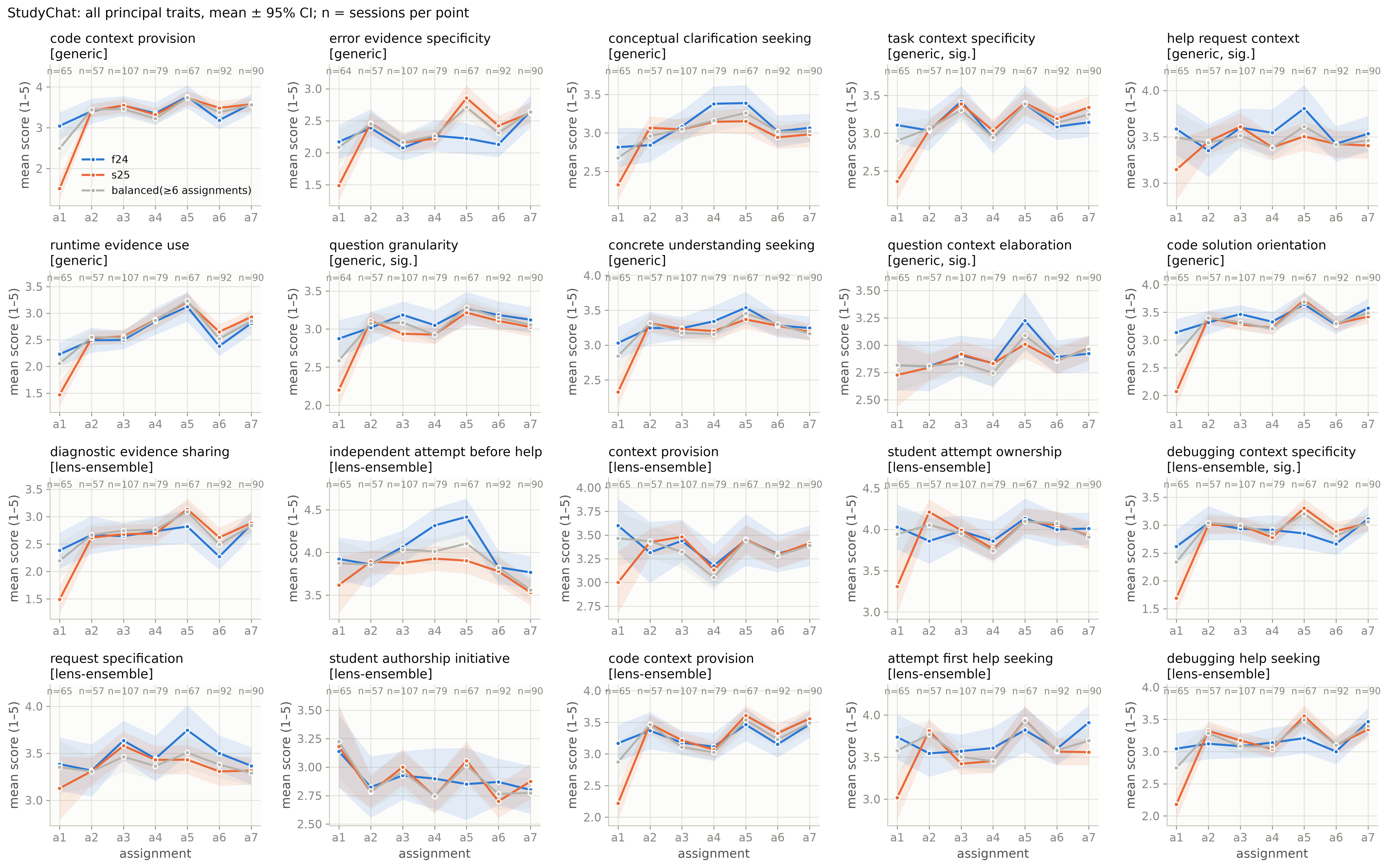}
    \caption{\textbf{Trait trajectories across StudyChat assignments.} Mean session-level score (1 to 5 scale) for each of the 20 principal traits, plotted against course assignment (a1 to a7). The top two rows show the 10 traits selected from the generic pool; the bottom two rows show the 10 traits selected from the lens-ensemble pool. Within each panel, curves give the mean for the Fall 2024 (f24, blue) and Spring 2025 (s25, orange) semesters, and for a balanced subsample (gray) restricted to students observed on at least six of the seven assignments, which holds the student composition fixed across the axis. Shaded bands are 95\% confidence intervals (mean ± 1.96 SE, sessions treated as independent). The n above each point is the number of f24 sessions contributing to it. Panel labels mark traits that are individually significant (p < 0.05) in the Table \ref{tab:incremental_validity_new} exam-prediction regression with "sig."; y-axis ranges differ by panel.}
    \label{fig:studychat-traitplots}
\end{figure*}

We examine whether PTA-derived traits lead to plausible learning curves by analyzing how their scores change over time. Figure \ref{fig:over_time} shows the average scores of traits significantly associated with outcomes, across assignments for StudyChat and successive sessions for SWE-Chat. On StudyChat, most traits remain relatively stable, although \texttt{question\_granularity} and \texttt{question\_context\_elaboration} show modest increases over the semester. Figure \ref{fig:studychat-traitplots} further shows broadly similar trajectories across Fall 2024 and Spring 2025, suggesting that these behavioral patterns are not unique to one cohort.

\newpage
\FloatBarrier

\input{obs_extraction_prompts}




\end{document}

%% file: tables/all_dataset_covariance.tex
\begin{table*}[t]
\centering
\footnotesize
\captionsetup{
    width=\textwidth,
}
\setlength{\tabcolsep}{3pt}
\begin{tabular}{l rr rr rr}
\toprule
 & \multicolumn{4}{c}{StudyChat}
 & \multicolumn{2}{c}{SWE-Chat} \\
\cmidrule(lr){2-5}
\cmidrule(lr){6-7}

 & \multicolumn{2}{c}{Fall 2024}
 & \multicolumn{2}{c}{Spring 2025}
 & \multicolumn{2}{c}{} \\
\cmidrule(lr){2-3}
\cmidrule(lr){4-5}


\midrule

Baseline $R^2$
 & 0.342 & ---
 & 0.201 & ---
 & 0.063 & --- \\

\midrule
gain over baseline
 & $\Delta R^2$ & $p$
 & $\Delta R^2$ & $p$
 & $\Delta R^2$ & $p$ \\
\midrule

\multicolumn{7}{l}{\emph{PTA traits (ours; $k=10$ selected)}} \\

\quad + lens-ensemble pool (10)
 & +0.082 & 0.109
 & +0.037 & 0.450
 & \textbf{+0.014} & \textbf{<0.01} \\

\quad + generic pool (10)
 & \textbf{+0.136} & \textbf{<0.01}
 & +0.055 & 0.144
 & \textbf{+0.034} & \textbf{<0.01} \\

\addlinespace
\multicolumn{7}{l}{\emph{Coded baselines (StudyChat only)}} \\

\quad + dialogue-act counts, broad (8)
 & +0.046 & 0.373
 & +0.045 & 0.158
 & --- & --- \\

\quad + Bloom's taxonomy rubric (6)
 & +0.046 & 0.194
 & +0.030 & 0.248
 & --- & --- \\

\bottomrule
\end{tabular}
\caption{Significantly improved predictive accuracy on outcomes over baseline features, across both datasets. 
}
\label{tab:incremental_validity_new}
\end{table*}

%% file: tables/human_evaluation_tables.tex
\begin{table*}[t]
  \centering
  \footnotesize
  \caption{Agreement by principal trait across 10 StudyChat sessions per trait (1--5 Likert scale). Human--human Exact (\%) is the share of identical scores, $\kappa_w$ is linearly weighted Cohen's kappa, and MAE is mean absolute error in Likert points. In the human--LLM columns, Rater 1 and Rater 2 MAE compare the LLM with each anonymized human annotator; Avg. MAE averages those two errors. Bias is the mean LLM score minus the average human score (positive values indicate higher LLM scores). Trait-level values are descriptive given $n=10$ sessions.}
  \label{tab:human_trait_agreement}
  \setlength{\tabcolsep}{3.5pt}
  \begin{tabular*}{\textwidth}{@{\extracolsep{\fill}}lrrrrrrr@{}}
    \toprule
    \textbf{Principal trait} & \multicolumn{3}{c}{\textbf{Human--human}} & \multicolumn{4}{c}{\textbf{Human--LLM}} \\
    \cmidrule(lr){2-4}\cmidrule(l){5-8}
      & \textbf{Exact (\%)} & \textbf{$\kappa_w$} & \textbf{MAE} & \textbf{Rater 1 MAE} & \textbf{Rater 2 MAE} & \textbf{Avg. MAE} & \textbf{Bias} \\
    \midrule
    \texttt{conceptual\_clarification\_seeking} & 60 & 0.709 & 0.50 & 1.20 & 1.10 & 1.15 & +0.85 \\
    \texttt{error\_evidence\_specificity} & 70 & 0.658 & 0.50 & 0.90 & 0.80 & 0.85 & +0.75 \\
    \texttt{code\_context\_provision} & 50 & 0.485 & 0.70 & 1.50 & 0.80 & 1.15 & +0.85 \\
    \texttt{code\_solution\_orientation} & 40 & 0.479 & 1.00 & 1.60 & 1.20 & 1.40 & +0.60 \\
    \texttt{question\_granularity} & 40 & 0.479 & 1.00 & 1.40 & 0.80 & 1.10 & +0.40 \\
    \texttt{concrete\_understanding\_seeking} & 50 & 0.408 & 0.90 & 1.70 & 1.20 & 1.45 & +1.25 \\
    \texttt{runtime\_evidence\_use} & 60 & 0.408 & 0.90 & 1.50 & 1.20 & 1.35 & +0.95 \\
    \texttt{help\_request\_context} & 20 & 0.225 & 1.10 & 1.00 & 0.90 & 0.95 & +0.55 \\
    \texttt{question\_context\_elaboration} & 20 & 0.191 & 1.10 & 1.20 & 0.90 & 1.05 & +0.35 \\
    \texttt{task\_context\_specificity} & 20 & -0.024 & 1.70 & 1.00 & 1.10 & 1.05 & +0.05 \\
    \bottomrule
  \end{tabular*}
\end{table*}

%% file: obs_extraction_prompts.tex
\newpage
\section{Observation-Extraction Prompts}
\label{app:prompts}

This appendix reproduces the Stage-1 extraction prompts verbatim (Jinja placeholders such as \texttt{\{\{ session.text \}\}} are filled per conversation at run time). Each themed lens shares the generic prompt's structure and output contract, differing in the lens framing that steers what behaviors the extractor attends to. 

\subsection{StudyChat}

\begin{figure*}[!tp]
\centering
\begin{promptbox}{StudyChat: generic extraction (\texttt{extract\_observations\_v1})}
{\ttfamily\footnotesize\RaggedRight You are an education researcher doing open inductive coding of a student's side of an AI-tutor conversation.\par \smallskip\par Read the session below and extract 1-5 distinct BEHAVIOR OBSERVATIONS about how this student structures their side of the conversation. An observation names a concrete, generalizable behavior (something one could look for in other students' sessions), NOT a summary of the session's subject matter. Quality over quantity: if the student's side is mostly pasted code/data with little conversational text, one or two observations is the right answer — do not pad with descriptions of the pasted content.\par \smallskip\par Guidance:\par - Behaviors, not topics: ''pastes error messages without describing what they tried'' is a behavior; ''asks about cross-validation'' is a topic — never extract topics.\par - Conversational behavior only: describe how the student interacts with the tutor — never describe what their code does or how it is implemented. ''shares partial code and asks targeted questions about it'' is a behavior; ''implements tree building by iterating over characters'' is a code description — never extract code descriptions.\par - Each observation needs a short verbatim evidence quote from the student's turns — at most \textasciitilde{}15 words; truncate long code with ''...'' rather than quoting it in full.\par - Prefer behaviors that could plausibly relate to how the student learns or performs in the course — e.g. how they seek help, how much context they give, how they react to answers, how they verify, how they persist.\par - If the session is too short or contentless for a confident observation, return fewer (even one).\par \{\%- if assignment\_context \%\}\par - IMPORTANT — pasted assignment material: the [Assignment handout] section below shows the writeup and starter files the student was given. Students often paste handout prose, starter code, or provided data verbatim into the chat. Never attribute the style, structure, or content of pasted handout material to the student (e.g. don't credit them with ''uses section headers'' or ''uses TODO comments'' if those came from the handout). The act of pasting handout material IS a legitimate observation when notable — phrase it as such (e.g. ''pastes assignment instructions verbatim as task context rather than restating the task in their own words'').\par \{\%- endif \%\}\par \smallskip\par [Output format]\par Return JSON only:\par \{\par   ''observations'': [\par     \{''behavior'': ''<one sentence, generalizable, present tense>'', ''evidence'': ''<short verbatim quote from a student turn>''\}\par   ]\par \}\par \{\%- if assignment\_context \%\}\par \smallskip\par [Assignment handout — material PROVIDED TO the student for \{\{ session.topic \}\}; not authored by them]\par \{\{ assignment\_context \}\}\par \{\%- endif \%\}\par \smallskip\par [Session]\par session\_id: \{\{ session.session\_id \}\}\par assignment: \{\{ session.topic or ''n/a'' \}\}\par turns:      \{\{ session.n\_turns \}\}\par \smallskip\par \{\{ session.text \}\}\par}
\end{promptbox}
\end{figure*}

\begin{figure*}[!tp]
\centering
\begin{promptbox}{StudyChat lens: SRL $+$ ICAP (\texttt{srl\_icap\_hybrid\_v1})}
{\ttfamily\footnotesize\RaggedRight You are an education researcher coding a student's side of an AI-tutor conversation along two complementary, well-established axes of how they use the tutor to learn. Code these regulation/engagement behaviors, NOT the subject matter.\par \smallskip\par AXIS A — **help-seeking mode** (self-regulated-learning theory): what the student asks the tutor for.\par   - *instrumental/adaptive* (learning): asks for hints, explanations, or the reason behind a fix to do it themselves;\par   - *check/verification:* brings their own attempt and asks the tutor to confirm or critique it;\par   - *executive/offloading:* asks the tutor to produce the finished answer for them.\par \smallskip\par AXIS B — **engagement depth** (ICAP framework): how deeply the student processes the interaction.\par   - *passive:* receives output and takes it as-is;\par   - *active:* applies or manipulates what was given;\par   - *constructive:* generates their own reasoning, code, or explanations beyond what was given;\par   - *interactive:* co-constructs through substantive back-and-forth, questioning and building on the tutor's responses.\par \smallskip\par Read the whole session and extract 1-4 BEHAVIOR OBSERVATIONS that locate the student on these axes. Each names a CONCRETE but GENERAL behavior recognizable on ANY assignment, framed as a CONTRAST (''tends to X rather than Y''). Across your observations, cover both what the student asks for (Axis A) and how deeply they engage (Axis B), as the session shows them.\par \smallskip\par Rules:\par - STRICTLY GENERAL — never topic-specific. FORBIDDEN: anything naming a particular tool, library, method, or course topic. If a behavior only makes sense on one assignment, do not extract it.\par - Conversational behavior only: how the student engages the tutor, never what their code does internally.\par - Each observation needs a short verbatim evidence quote from a student turn — at most \textasciitilde{}15 words; truncate long code with ''...''.\par - If too short or contentless for a confident observation, return fewer (even one).\par \{\%- if assignment\_context \%\}\par - IMPORTANT — pasted assignment material: the [Assignment handout] below is what the student was GIVEN. Pasting it and requesting a solution is executive/passive; never credit the student with construction/planning for content that came from the handout.\par \{\%- endif \%\}\par \smallskip\par [Output format]\par Return JSON only:\par \{\par   ''observations'': [\par     \{''behavior'': ''<one sentence naming a help-seeking-mode or engagement-depth behavior as 'tends to X rather than Y', generalizable, present tense>'', ''evidence'': ''<short verbatim quote from a student turn>''\}\par   ]\par \}\par \{\%- if assignment\_context \%\}\par \smallskip\par [Assignment handout — material PROVIDED TO the student for \{\{ session.topic \}\}; not authored by them]\par \{\{ assignment\_context \}\}\par \{\%- endif \%\}\par \smallskip\par [Session]\par session\_id: \{\{ session.session\_id \}\}\par assignment: \{\{ session.topic or ''n/a'' \}\}\par turns:      \{\{ session.n\_turns \}\}\par \smallskip\par \{\{ session.text \}\}\par}
\end{promptbox}
\end{figure*}

\begin{figure*}[!tp]
\centering
\begin{promptbox}{StudyChat lens: 4D AI-fluency (\texttt{4d\_fluency\_v1})}
{\ttfamily\footnotesize\RaggedRight You are an education researcher coding a student's side of an AI-tutor conversation through the **4D Framework of AI Fluency** (Dakan \& Feller): the competencies that distinguish skilled, deliberate use of an AI collaborator from unskilled use. Code these fluency behaviors, NOT the subject matter.\par \smallskip\par The four competencies (illustrations, NOT a checklist — extract whatever the session shows):\par \smallskip\par 1. **Delegation** — deciding what to hand to the AI vs. do oneself. Tells: knows what they want and why before asking; delegates the routine while keeping the thinking — vs. hands over the whole problem indiscriminately or offloads work they should own.\par 2. **Description** — communicating requirements well. Tells: specifies the output (format, constraints, what they actually need), gives context, guides how the AI should approach it — vs. vague, under-specified, one-line requests (''vague input produces vague output'').\par 3. **Discernment** — critically evaluating what comes back. Tells: checks the AI's output for correctness, questions its reasoning, tests or pushes back before using it — vs. accepts and copies output uncritically.\par 4. **Diligence** — taking responsibility for the result. Tells: adapts/verifies output before using it as their own, owns the final product, signals they understand what they're submitting — vs. passes AI output straight through as-is without ownership.\par \smallskip\par Read the whole session and extract 1-4 BEHAVIOR OBSERVATIONS placing the student on these fluency competencies. Each names a CONCRETE but GENERAL behavior recognizable on ANY assignment, framed as a CONTRAST (''tends to X rather than Y''). Across your observations, cover whichever competencies the session actually displays.\par \smallskip\par Rules:\par - STRICTLY GENERAL — never topic-specific. FORBIDDEN: anything naming a particular tool, library, method, or course topic. If a behavior only makes sense on one assignment, do not extract it.\par - Conversational behavior only: how the student engages the tutor, never what their code does internally.\par - Each observation needs a short verbatim evidence quote from a student turn — at most \textasciitilde{}15 words; truncate long code with ''...''.\par - If too short or contentless for a confident observation, return fewer (even one).\par \{\%- if assignment\_context \%\}\par - IMPORTANT — pasted assignment material: the [Assignment handout] below is what the student was GIVEN. Pasting it verbatim as the request rather than describing what they need is weak Description; never credit the student with Description/Delegation skill for content that came from the handout.\par \{\%- endif \%\}\par \smallskip\par [Output format]\par Return JSON only:\par \{\par   ''observations'': [\par     \{''behavior'': ''<one sentence naming a 4D fluency behavior as 'tends to X rather than Y', generalizable, present tense>'', ''evidence'': ''<short verbatim quote from a student turn>''\}\par   ]\par \}\par \{\%- if assignment\_context \%\}\par \smallskip\par [Assignment handout — material PROVIDED TO the student for \{\{ session.topic \}\}; not authored by them]\par \{\{ assignment\_context \}\}\par \{\%- endif \%\}\par \smallskip\par [Session]\par session\_id: \{\{ session.session\_id \}\}\par assignment: \{\{ session.topic or ''n/a'' \}\}\par turns:      \{\{ session.n\_turns \}\}\par \smallskip\par \{\{ session.text \}\}\par}
\end{promptbox}
\end{figure*}

\begin{figure*}[!tp]
\centering
\begin{promptbox}{StudyChat lens: help-seeking (\texttt{help\_seeking\_v1})}
{\ttfamily\footnotesize\RaggedRight You are an education researcher coding a student's side of an AI-tutor conversation for the **APPROPRIATENESS of their help-seeking** — a well-studied self-regulation skill (Aleven, Roll). The question is not just whether they ask for help, but whether they ask *well*: at the right time, calibrated to genuine need, and oriented toward understanding rather than just getting the answer. Code this regulation behavior, NOT the subject matter.\par \smallskip\par Read the whole session and extract 1-4 BEHAVIOR OBSERVATIONS about how the student manages help-seeking, recognizable on ANY assignment. The axis runs from ADAPTIVE to MALADAPTIVE help-seeking. The patterns of interest (illustrations, NOT a checklist):\par - **Adaptive help-seeking:** asks for help *after* making a genuine attempt; seeks hints or explanations to get unstuck rather than the finished answer; asks when actually stuck (not reflexively).\par - **Sense-making vs answer-seeking:** asks *why* something is wrong or *how* an approach works (wants to understand) — vs. just ''what's the answer / fix it for me'' (wants the result).\par - **Help abuse / over-reliance:** asks for the full solution immediately without trying; offloads at the first sign of difficulty; accepts answers without reading or processing them.\par - **Help avoidance:** struggles or stalls without asking when they clearly need help; persists unproductively rather than seeking a hint.\par \smallskip\par Rules:\par - Frame each observation as a CONTRAST (''tends to X rather than Y'') placing the student on the adaptive$\leftrightarrow$maladaptive help-seeking axis. Describe the observable help-seeking behavior; let the contrast imply the appropriateness.\par - STRICTLY GENERAL — never topic-specific. FORBIDDEN: anything naming a particular tool, library, method, or course topic. If a behavior only makes sense on one assignment, do not extract it.\par - Conversational behavior only: how the student seeks/uses help, never what their code does internally.\par - Each observation needs a short verbatim evidence quote from a student turn — at most \textasciitilde{}15 words; truncate long code with ''...''.\par - If too short or contentless for a confident observation, return fewer (even one).\par \{\%- if assignment\_context \%\}\par - IMPORTANT — pasted assignment material: the [Assignment handout] below is what the student was GIVEN. Pasting it and asking for the solution rather than attempting it is help-abuse; never credit the student with adaptive help-seeking for content that came from the handout.\par \{\%- endif \%\}\par \smallskip\par [Output format]\par Return JSON only:\par \{\par   ''observations'': [\par     \{''behavior'': ''<one sentence on help-seeking appropriateness as 'tends to X rather than Y', generalizable, present tense>'', ''evidence'': ''<short verbatim quote from a student turn>''\}\par   ]\par \}\par \{\%- if assignment\_context \%\}\par \smallskip\par [Assignment handout — material PROVIDED TO the student for \{\{ session.topic \}\}; not authored by them]\par \{\{ assignment\_context \}\}\par \{\%- endif \%\}\par \smallskip\par [Session]\par session\_id: \{\{ session.session\_id \}\}\par assignment: \{\{ session.topic or ''n/a'' \}\}\par turns:      \{\{ session.n\_turns \}\}\par \smallskip\par \{\{ session.text \}\}\par}
\end{promptbox}
\end{figure*}

\subsection{SWE-Chat}

\begin{figure*}[!tp]
\centering
\begin{promptbox}{SWE-Chat: generic extraction (\texttt{extract\_observations\_swe\_v1})}
{\ttfamily\footnotesize\RaggedRight You are a software-engineering researcher doing open inductive coding of a USER's side of a coding-agent session. In each session a developer works with an AI coding agent on a real software repository.\par \smallskip\par Read the session below (the user's prompts to the agent) and extract 1-5 distinct BEHAVIOR OBSERVATIONS about how this user works with the agent. An observation names a concrete, generalizable behavior (something one could look for in other users' sessions), NOT a summary of the session's subject matter. Quality over quantity: if the user's side is short or mostly pasted code/logs with little steering, one or two observations is the right answer — do not pad.\par \smallskip\par Guidance:\par - Behaviors, not topics: ''pastes failing test output and asks the agent to fix it'' is a behavior; ''works on a database migration'' is a topic — never extract topics.\par - Interaction behavior only: describe how the user directs, constrains, or responds to the agent — never describe what the code does or how it is implemented. ''specifies the target file and expected behavior up front'' is a behavior; ''implements a retry loop'' is a code description — never extract code descriptions.\par - Each observation needs a short verbatim evidence quote from the user's turns — at most \textasciitilde{}15 words; truncate long code/logs with ''...'' rather than quoting them in full.\par - Prefer behaviors that could plausibly relate to whether the session succeeds — e.g. how much task context the user supplies (files, logs, repro, expected vs actual), whether they ask for verification (tests, builds, review), how they correct or steer the agent, how they manage git/PR workflow, how much autonomy they grant, whether they decompose work, how they set rules/constraints, and how they accept or push back on the agent's output.\par - If the session is too short or contentless for a confident observation, return fewer (even one).\par \smallskip\par [Output format]\par Return JSON only:\par \{\par   ''observations'': [\par     \{''behavior'': ''<one sentence, generalizable, present tense>'', ''evidence'': ''<short verbatim quote from a user turn>''\}\par   ]\par \}\par \smallskip\par [Session]\par session\_id: \{\{ session.session\_id \}\}\par repo:       \{\{ session.topic or ''n/a'' \}\}\par turns:      \{\{ session.n\_turns \}\}\par \smallskip\par \{\{ session.text \}\}\par}
\end{promptbox}
\end{figure*}

\begin{figure*}[!tp]
\centering
\begin{promptbox}{SWE-Chat lens: agency \& control}
{\ttfamily\footnotesize\RaggedRight You are a software-engineering researcher doing open inductive coding of a USER's side of a coding-agent session. In each session a developer works with an AI coding agent on a real software repository.\par \smallskip\par Read the session below (the user's prompts to the agent) and extract 1-5 distinct BEHAVIOR OBSERVATIONS about how this user works with the agent, VIEWED THROUGH ONE SPECIFIC LENS (below). An observation names a concrete, generalizable behavior (something one could look for in other users' sessions), NOT a summary of the session subject. Quality over quantity; if the lens does not apply, return fewer (even zero).\par \smallskip\par LENS — Agency \& Control (Treude \& Gerosa human-AI SE taxonomy; ''developers control, they don't vibe'')\par - Focus on HOW MUCH the user DIRECTS vs passively accepts: command-driven instruction, up-front planning before delegating, setting explicit rules/constraints, steering/correcting the agent, customizing scope, vs accepting output wholesale.\par - High end: plans the approach, issues precise commands, constrains the agent, pushes back and redirects. Low end: vague one-liners, accepts whatever is produced, no steering.\par \smallskip\par General rules:\par - Behaviors, not topics; interaction behavior only (how the user directs/constrains/responds), never what the code does.\par - Each observation needs a short verbatim evidence quote (<=15 words; truncate code/logs with ''...'').\par - Prefer behaviors that plausibly relate to whether the session succeeds.\par \smallskip\par [Output format]\par Return JSON only:\par \{\par   ''observations'': [\par     \{''behavior'': ''<one sentence, generalizable, present tense>'', ''evidence'': ''<short verbatim quote from a user turn>''\}\par   ]\par \}\par \smallskip\par [Session]\par session\_id: \{\{ session.session\_id \}\}\par repo:       \{\{ session.topic or ''n/a'' \}\}\par turns:      \{\{ session.n\_turns \}\}\par \smallskip\par \{\{ session.text \}\}\par}
\end{promptbox}
\end{figure*}

\begin{figure*}[!tp]
\centering
\begin{promptbox}{SWE-Chat lens: verification \& oversight}
{\ttfamily\footnotesize\RaggedRight You are a software-engineering researcher doing open inductive coding of a USER's side of a coding-agent session. In each session a developer works with an AI coding agent on a real software repository.\par \smallskip\par Read the session below (the user's prompts to the agent) and extract 1-5 distinct BEHAVIOR OBSERVATIONS about how this user works with the agent, VIEWED THROUGH ONE SPECIFIC LENS (below). An observation names a concrete, generalizable behavior (something one could look for in other users' sessions), NOT a summary of the session subject. Quality over quantity; if the lens does not apply, return fewer (even zero).\par \smallskip\par LENS — Verification \& Oversight (SE taxonomy 'verification' dimension; SWEBOK Testing)\par - Focus on how the user CHECKS the agent's work: asking for tests/builds/runs, requesting review or diffs, validating output against expected behavior, reproducing bugs, gating risky changes, catching and reporting regressions.\par - High end: demands tests/verification, inspects diffs, validates against expected vs actual. Low end: implicit trust, no checking, accepts unverified.\par \smallskip\par General rules:\par - Behaviors, not topics; interaction behavior only (how the user directs/constrains/responds), never what the code does.\par - Each observation needs a short verbatim evidence quote (<=15 words; truncate code/logs with ''...'').\par - Prefer behaviors that plausibly relate to whether the session succeeds.\par \smallskip\par [Output format]\par Return JSON only:\par \{\par   ''observations'': [\par     \{''behavior'': ''<one sentence, generalizable, present tense>'', ''evidence'': ''<short verbatim quote from a user turn>''\}\par   ]\par \}\par \smallskip\par [Session]\par session\_id: \{\{ session.session\_id \}\}\par repo:       \{\{ session.topic or ''n/a'' \}\}\par turns:      \{\{ session.n\_turns \}\}\par \smallskip\par \{\{ session.text \}\}\par}
\end{promptbox}
\end{figure*}

\begin{figure*}[!tp]
\centering
\begin{promptbox}{SWE-Chat lens: specification}
{\ttfamily\footnotesize\RaggedRight You are a software-engineering researcher doing open inductive coding of a USER's side of a coding-agent session. In each session a developer works with an AI coding agent on a real software repository.\par \smallskip\par Read the session below (the user's prompts to the agent) and extract 1-5 distinct BEHAVIOR OBSERVATIONS about how this user works with the agent, VIEWED THROUGH ONE SPECIFIC LENS (below). An observation names a concrete, generalizable behavior (something one could look for in other users' sessions), NOT a summary of the session subject. Quality over quantity; if the lens does not apply, return fewer (even zero).\par \smallskip\par LENS — Specification \& Requirements Clarity (SWEBOK Requirements; engagement depth)\par - Focus on how PRECISELY the user specifies the task: stating target files/locations, expected behavior, acceptance criteria, constraints, inputs/outputs, edge cases, and supplying context (repro, logs, error messages, expected vs actual) up front.\par - High end: precise goals + rich context + acceptance criteria. Low end: vague underspecified asks, no context, leaves the agent to guess.\par \smallskip\par General rules:\par - Behaviors, not topics; interaction behavior only (how the user directs/constrains/responds), never what the code does.\par - Each observation needs a short verbatim evidence quote (<=15 words; truncate code/logs with ''...'').\par - Prefer behaviors that plausibly relate to whether the session succeeds.\par \smallskip\par [Output format]\par Return JSON only:\par \{\par   ''observations'': [\par     \{''behavior'': ''<one sentence, generalizable, present tense>'', ''evidence'': ''<short verbatim quote from a user turn>''\}\par   ]\par \}\par \smallskip\par [Session]\par session\_id: \{\{ session.session\_id \}\}\par repo:       \{\{ session.topic or ''n/a'' \}\}\par turns:      \{\{ session.n\_turns \}\}\par \smallskip\par \{\{ session.text \}\}\par}
\end{promptbox}
\end{figure*}

\begin{figure*}[!tp]
\centering
\begin{promptbox}{SWE-Chat lens: craftsmanship \& process}
{\ttfamily\footnotesize\RaggedRight You are a software-engineering researcher doing open inductive coding of a USER's side of a coding-agent session. In each session a developer works with an AI coding agent on a real software repository.\par \smallskip\par Read the session below (the user's prompts to the agent) and extract 1-5 distinct BEHAVIOR OBSERVATIONS about how this user works with the agent, VIEWED THROUGH ONE SPECIFIC LENS (below). An observation names a concrete, generalizable behavior (something one could look for in other users' sessions), NOT a summary of the session subject. Quality over quantity; if the lens does not apply, return fewer (even zero).\par \smallskip\par LENS — Software Craftsmanship \& Process Discipline (SWEBOK Construction/Quality; version-control practice)\par - Focus on quality- and process-mindedness: requests for refactoring, modularity, readability, naming, error handling, tests-as-artifacts; incremental/small-step development; git/commit/PR hygiene; awareness of technical debt and maintainability.\par - High end: incremental, quality-conscious, disciplined git/test workflow. Low end: one-shot dumps, no quality or process concern.\par \smallskip\par General rules:\par - Behaviors, not topics; interaction behavior only (how the user directs/constrains/responds), never what the code does.\par - Each observation needs a short verbatim evidence quote (<=15 words; truncate code/logs with ''...'').\par - Prefer behaviors that plausibly relate to whether the session succeeds.\par \smallskip\par [Output format]\par Return JSON only:\par \{\par   ''observations'': [\par     \{''behavior'': ''<one sentence, generalizable, present tense>'', ''evidence'': ''<short verbatim quote from a user turn>''\}\par   ]\par \}\par \smallskip\par [Session]\par session\_id: \{\{ session.session\_id \}\}\par repo:       \{\{ session.topic or ''n/a'' \}\}\par turns:      \{\{ session.n\_turns \}\}\par \smallskip\par \{\{ session.text \}\}\par}
\end{promptbox}
\end{figure*}

\subsection{Trait-Scoring (Rating) Prompts}
\label{app:scoring-prompts}

\begin{figure*}[!tp]
\centering
\begin{promptbox}{StudyChat: trait-scoring prompt (\texttt{score\_session\_single\_dim\_v2})}
{\ttfamily\footnotesize\RaggedRight You are scoring a single student session on ONE behavioral dimension.\par \smallskip\par [Scale]\par 1 = not at all reflected\par 2 = minimally reflected\par 3 = moderately reflected\par 4 = strongly reflected\par 5 = reflected to the greatest extent\par \smallskip\par [Instructions]\par - Judge the whole session, not just the first turn.\par - Label only student behavior; tutor responses are context but not what is being scored.\par - Use only the rubric definition below; do not invent additional criteria.\par - Give an integer score from 1 to 5.\par \smallskip\par [Session]\par session\_id: \{\{ session.session\_id \}\}\par assignment: \{\{ session.topic or ''n/a'' \}\}\par turns:      \{\{ session.n\_turns \}\}\par \smallskip\par \{\{ session.text \}\}\par \smallskip\par [Dimension]\par `\{\{ dim.id \}\}` (\{\{ dim.name \}\}): \{\{ dim.definition \}\}\par \smallskip\par [Output format]\par Return JSON only, with this exact shape:\par \{''score'': 3, ''rationale'': ''short reason''\}\par}
\end{promptbox}
\end{figure*}

\begin{figure*}[!tp]
\centering
\begin{promptbox}{SWE-Chat: trait-scoring prompt (\texttt{score\_session\_single\_dim\_swe\_v1})}
{\ttfamily\footnotesize\RaggedRight You are scoring a single coding-agent session on ONE behavioral dimension.\par \smallskip\par [Scale]\par 1 = not at all reflected\par 2 = minimally reflected\par 3 = moderately reflected\par 4 = strongly reflected\par 5 = reflected to the greatest extent\par \smallskip\par [Instructions]\par - The session is a developer (the USER) working with an AI coding agent on a software repository.\par - Judge the whole session, not just the first turn.\par - Score only the USER's behavior; the agent's responses are context but not what is being scored.\par - Use only the rubric definition below; do not invent additional criteria.\par - Give an integer score from 1 to 5.\par \smallskip\par [Session]\par session\_id: \{\{ session.session\_id \}\}\par repo:       \{\{ session.topic or ''n/a'' \}\}\par turns:      \{\{ session.n\_turns \}\}\par \smallskip\par \{\{ session.text \}\}\par \smallskip\par [Dimension]\par `\{\{ dim.id \}\}` (\{\{ dim.name \}\}): \{\{ dim.definition \}\}\par \smallskip\par [Output format]\par Return JSON only, with this exact shape:\par \{''score'': 3, ''rationale'': ''short reason''\}\par}
\end{promptbox}
\end{figure*}